\documentclass[preprint,review,number,12pt]{elsarticle}
\usepackage[margin=3cm]{geometry}
\usepackage{graphicx}
\usepackage{color}
\usepackage{subfigure}
\usepackage{textcomp}
\usepackage{amssymb}

\newcommand{\pd}[2]{\frac{\partial #1}{\partial #2}}

\graphicspath{{figs/}{eps/}}

\biboptions{square,sort}

\journal{???}

\begin{document}

\begin{frontmatter}

\title{Simultaneous Feature and Expert Selection  \\  within Mixture of Experts}

\author[rvt]{Billy Peralta\corref{cor1}}
\ead{bperalta@uct.cl}
\cortext[cor1]{Corresponding author, Telephone: (56 45) 255 3948}
\address[rvt]{Department of Informatics,Universidad Cat\'olica de Temuco, Chile.}


\begin{abstract}

A useful strategy to deal with complex classification scenarios is the ``divide
and conquer'' approach. The mixture of experts (MOE) technique makes use of this
strategy by joinly training a set of classifiers, or experts, that
are specialized in different regions of the input space. A global model, or gate
function, complements the experts by learning a function that weights their
relevance in different parts of the input space. Local feature selection appears
as an attractive alternative to
improve the specialization of experts and gate
function, particularly, for the case of high dimensional data. Our
main intuition is that particular subsets of
dimensions, or subspaces, are usually more appropriate to classify
instances located in different regions of the input space. Accordingly, this
work contributes with a
regularized variant of MoE that incorporates an embedded process for
local feature selection using $L1$ regularization, with a simultaneous expert selection. The experiments are still pending.

\end{abstract}
\begin{keyword}
Mixture of experts, local feature selection, embedded feature
selection, regularization.
\end{keyword}

\end{frontmatter}

\section{Mixture of Experts with embedded variable selection} \label{Sec:OurApproach}

Our main idea is to incorporate a local feature selection scheme inside each
expert and gate function of a MoE formulation. Our main intuition is that,
in the context of classification, different partitions of the input data can be
best represented by specific subsets of features. This is particularly relevant
in the case of high dimensional spaces, where the common presence of noisy or
irrelevant features might obscure the detection of particular class patterns.
Specifically, our approach takes advantage of the linear nature of each local
expert and gate function in the classical MoE formulation \cite{1351018},
meaning that $L1$ regularization can be directly applied. Below, we first
briefly describe the classical MoE formulation for classification. Afterwards,
we discuss the proposed modification to the MoE model that provides embedded
feature selection. 

\subsection{Mixture of Experts}
In the context of supervised classification, there is available a set of $N$
training examples, or instance-label pairs $(x_n,y_n)$,
representative of the domain data $(x,y)$, where $x_n \in \Re^D$ and $y_n \in
C$. Here $C$ is a discrete set
of $Q$ class labels $\left\{c_1,...,c_Q\right\}$. The goal is to use 
training data to find a function $f$ that minimizes a loss function which
scores the quality of $f$ to predict the true underlying relation between $x$
and $y$.
From a probabilistic point of view \cite{Bishop:2007}, a useful
approach to find $f$ is using a conditional formulation: 

\begin{eqnarray}
f(x) & = & \arg\max_{c_i \in  C } \: p(y=c_i | x) \label{eq:1}.
\nonumber
\end{eqnarray}

In the general case of complex relations between $x$ and $y$, a useful
strategy consists of approximating $f$ through a mixture of local functions.
This
is similar to the case of modeling a mixture distribution
\cite{citeulike:2235458} and it leads to the MoE model.

\begin{figure}[h!] 
\centering
\includegraphics[trim = 20mm 180mm 40mm 20mm, clip, width=300pt]{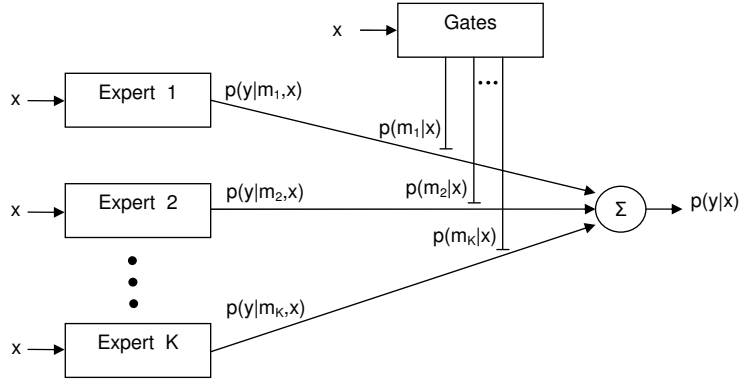}
\caption{Mixture of experts scheme.}
\label{fig:Fig_14}
\end{figure}

We decompose the conditional likelihood $p(y|x)$ as:
\begin{eqnarray}\label{Eq:MoE}
p(y|x) & = & \sum^{K}_{i=1} p(y,m_i | x) \;=\; \sum^{K}_{i=1} p(y |
m_i,x)\:p(m_i | x), 
\end{eqnarray}

\noindent where Equation (\ref{Eq:MoE}) represents a MoE model with $K$ experts
$m_i$. Figure (\ref{fig:Fig_14}) shows a schematic diagram
of
the
MoE approach. The main idea is to obtain local models in such a way that they
are
specialized in a
particular region of the data. In Figure (\ref{fig:Fig_14}), $x$ corresponds to
the
input instance, $p(y | m_i,x)$ is the \textbf{expert function}, $p(m_i | x)$ is the
\textbf{gating function}, and $p(y | x)$ is the weighted sum of the experts. Note that
the output of each expert model is weighted by the gating function. This weight
can be interpreted as the \textit{relevance} of expert $m_i$ for the
classification of input instance $x$. Also note that the gate function has $K$ outputs, one for
each expert. There are $K$ expert functions that have $Q$ components, one for each class.

%
%
%

The traditional MoE technique uses multinomial logit models, also
known as soft-max
functions \cite{Bishop:2007}, to represent the gate and expert functions. An
important characteristic of this model is that it forces competition among its
components. In MoE, such components are expert functions for the gates and
class-conditional functions for the experts. The competition in soft-max
functions enforces the especialization of experts in different areas of the
input space \cite{Yuille:1998:WM:303568.304791}. 

Using multinomial logit models, a gate function is defined as:

\begin{eqnarray} \label{Eq:MoE-Params-1}
p(m_i|x) & = &  \frac{exp{(\nu^{T}_{i}x)}}{\sum^{K}_{j=1} exp{(\nu^{T}_{j}x})} 
\end{eqnarray}


\noindent where $i \in \{1, \dots, K\}$ refers to the set of experts and
$\nu_i \in
\Re^D$ is a vector of model parameters. Component $\nu_{ij}$ of vector $\nu_i$
models the relation between the gate and dimension $j$ of input instance $x$.  

Similarly, an expert function is defined as:

\begin{eqnarray}\label{Eq:MoE-Params-2}
p(y=c_l|x,m_i) & = &  \frac{exp{(\omega^{T}_{li}x)}}{\sum^{M}_{j=1}
exp{(\omega^{T}_{ji}x)}} 
\end{eqnarray}


\noindent where $\omega_{li}$ depends on class label $c_l$ and expert $i$. In
this way, there are a total of $Q \times K$ vectors $\omega_{li}$. Component
$\omega_{lij}$ of
vector $\omega_{li}$ models the relation between expert function $i$ and
dimension $j$ of input instance $x$. 


There are several methods to find the value of the hidden parameters $\nu_{ij}$
and $\omega_{lij}$ \cite{Moerland97somemethods}. An attractive alternative is
to use the EM algorithm. In the case of MoE, the EM formulation augments the
model by introducing a set of latent variables, or \textit{responsibilities}, 
indicating the expert that generates each instance. Accordingly, the EM
iterations consider an expectation step that estimates expected values for
\textit{responsibilities}, and a maximization step that updates the values
of parameters $\nu_{ij}$
and $\omega_{lij}$. Specifically, the
posterior probability of the
\textit{responsibility} $R_{in}$ assigned by the gate function to expert
$m_i$ for an instance $x_n$ is given by \cite{Moerland97somemethods}:

\begin{eqnarray} \label{Eq:resp}
R_{in} & = & p(m_i|x_n,y_n) \\ 
       & = & \frac{p(y_n|x_n,m_i)\:p(m_i|x_n)}{\sum^{K}_{j=1} p(y_n |
x_n,m_j)\:p(m_j | x_n)}  \nonumber 
\end{eqnarray}

Considering these responsibilities and Equation (\ref{Eq:MoE}), the
expected complete log-likelihood $\left\langle
\textsl{L}_c \right\rangle$ used in the EM iterations is
\cite{Moerland97somemethods}:

\begin{eqnarray}\label{Eq:ExpLogL}
	\left\langle \textsl{L}_c \right\rangle & = & \sum^{N}_{n=1}{
\sum^{K}_{i=1}{ R_{in}\: \left[log \; p(y_n|x_n,m_i)\: + log \; p(m_i|x_n)
\right]}}
\end{eqnarray}

\subsection{Regularized Mixture of Experts (RMoE)}
To embed a feature selection process in the MoE approach, we use the fact that
in Equations (\ref{Eq:MoE-Params-1}) and (\ref{Eq:MoE-Params-2}) the multinomial
logit models for gate and experts functions contain linear relations for the
relevant parameters. This linearity can be straightforwardly used in feature
selection by considering that a parameter 
component $\nu_{ij}$ or $\omega_{lij}$ with zero value implies that dimension
${j}$ is irrelevant for gate function $p(m_i|x)$ or expert model $p(y
| m_i,x)$,
respectively. Consequently, we propose to penalize complex
models using $L_1$ regularization. Similar consideration is used in the work of
\cite{1248698} but in the context of unsupervised learning. The
idea is to maximize the likelihood of data
while simultaneously minimizing the number of parameter components $\nu_{ij}$
and
$\omega_{lij}$ different from zero. Considering that there are
$Q$ classes, $K$
experts, and $D$ dimensions, the expected $L1$ regularized log-likelihood
$\left\langle \textsl{L}^R_c \right\rangle$ is given by:

\begin{eqnarray}\label{Eq:ExpLogL-Reg}
    \left\langle \textsl{L}^R_c \right\rangle & = & \left\langle \textsl{L}_c
\right\rangle - \lambda_\nu \sum^{K}_{i=1}{ \sum^{D}_{j=1}{ \left| \nu_{ij}
\right| }} - \lambda_\omega \sum^{Q}_{l=1}{ \sum^{K}_{i=1}{ \sum^{D}_{j=1}{
\left| \omega_{lij} \right| }}} \; .
\end{eqnarray}

%

To maximize Equation (\ref{Eq:ExpLogL-Reg}) with respect to model parameters, we
use first the standard fact that the likelihood function in Equation
(\ref{Eq:ExpLogL}) can be decomposed in terms of independent expressions for
gate
and expert models \cite{Moerland97somemethods}. In this way, the
maximization step of the EM based solution can be performed independently with
respect to gate and expert parameters \cite{Moerland97somemethods}. In our
problem, each of these optimizations has an extra term given by the respective
regularization term in Equation (\ref{Eq:ExpLogL-Reg}). To handle this case, we
observe that each of these optimizations is equivalent to the expression to
solve a regularized logistic regression \cite{Lee+etal06:L1logreg}.
As shown in \cite{Lee+etal06:L1logreg}, this problem can be solved by using a
coordinate ascent optimization strategy \cite{Tseng:01} given by a sequential
two-step approach that first models the problem as an unregularized
logistic regression and afterwards incorporates the regularization
constraints. 

In summary, we handle Equation (\ref{Eq:ExpLogL-Reg}) by using a EM based
strategy that at each step solves the maximation with respect to model
parameters by decomposing this problem in terms of gate and expert parameters.
Each of these problems is in turn solved using the strategy proposed in
\cite{Lee+etal06:L1logreg}. Next, we provide details of this procedure.


\textbf{Optimization of the unregularized log-likelihood}

In this case, we solve the unconstrained log-likelihood given by Equation (\ref{Eq:ExpLogL}). First, we 
optimize the log-likelihood with respect to vector $\omega_{li}$. The maximization of the expected log-likelihood $\left\langle \textsl{L}_c
\right\rangle$ implies deriving Equation (\ref{Eq:ExpLogL}) with respect to
$\omega_{li}$:



%
%


\begin{eqnarray}\label{Eq:ExpLogL-Der-1}
\pd{ \sum^{N}_{n=1}{ \sum^{K}_{i=1}{ R_{in}\: \left[log \; p(y_n|x_n,m_i)\:
\right]}}}{\omega_{li}} & = & 0,     
\end{eqnarray}

\noindent and applying the derivate, we have:

\begin{eqnarray}\label{Eq:ExpLogL-Der-2}
- \sum^{N}_{n=1}{R_{in} \left( p(y_n|x_n,m_i) - y_n \right)x_n } & = & 0.
\end{eqnarray}

In this case, the classical technique of least-squares can not be
directly applied
because of the soft-max function in $p(y_n|x_n,m_i)$. Fortunately, as
described in \cite{journals/neco/JordanJ94} and later in
\cite{Moerland97somemethods}, Equation (\ref{Eq:ExpLogL-Der-2}) can be
approximated by
using a transformation that implies inverting the soft-max function. Using this
transformation, Equation (\ref{Eq:ExpLogL-Der-2}) is
equivalent to an optimization problem that can be solved using a weighted least 
squares technique \cite{Bishop:2007}:

\begin{eqnarray}
    \min_{\omega_{li}} & \sum^{N}_{n=1}{R_{in} \left( 
\omega^{T}_{li}x_n - log \: y_{n} \right )^2 } \label{eq:14}
 \end{eqnarray}

A similar derivation can be performed with respect to vectors $\nu_{i}$. Again
deriving Equation (\ref{Eq:ExpLogL}), in this case with respect to
parameters $\nu_{ij}$ and
applying the transformation suggested in \cite{journals/neco/JordanJ94}, we
obtain:

\begin{eqnarray}
    \min_{\nu_{i}} & \sum^{N}_{n=1}{\left(\nu^{T}_{i} x_n - log R_{in}
\right)^2 } \label{eq:14-b}\\
\end{eqnarray}

\textbf{Optimization of the regularized likelihood}

Following the procedure of \cite{Lee+etal06:L1logreg}, we add the regularization
term to the optimization problem given by Equation (\ref{eq:14}), obtaining an
expression that can be
solved using quadratic programming \cite{tibshirani96regression}:

\begin{eqnarray}
    \min_{\omega_{li}} & \sum^{N}_{n=1}{R_{in} \left( log \: y_{n} - \omega^{T}_{li}x_n \right)^2 } \nonumber \\
    \mbox{subject to:}    & ||\omega_{li}||_1 \leq \lambda_\omega
     \label{eq:15}
\end{eqnarray}

%

Similarly, we can
also obtain a standard quadratic
optimization problem to find parameters $\nu_{ij}$ :

\begin{eqnarray}
    \min_{\nu_{i}} & \sum^{N}_{n=1}{\left( log R_{in} - \nu^{T}_{i} x_n
\right)^2 } \nonumber\\
\mbox{subject: to}    & ||\nu_{i}||_1 \leq \lambda_\nu
     \label{eq:16}
\end{eqnarray}

A practical advantage of using quadratic programming is that most available
optimization packages can be utilized to solve it \cite{Boyd&Vandenberghe:2004}.
Specifically, in the case of $T$ iterations, there are a total of $T*K*(Q+1)$
convex quadratic problems
related to the maximization step of the EM algorithm. To further reduce this
computational load, we slightly modify this maximization by applying the following
two-steps scheme:

\begin{itemize} 
\item Step-1: Solve $K$ quadratic problems to find gate parameters $\nu_{ij}$
assuming that each expert uses all the available dimensions. In this case, there
are $T-1$ iterations. 
\item Step-2: Solve $K*(Q+1)$ quadratic problems to find expert parameters
$\omega_{lij}$ applying the feature selection process. In this case, there is a
single iteration.
\end{itemize}

Using the previous scheme we reduce from $T*K*(Q+1)$ to $K*(T+1)+K*(Q+1)$ the
number of quadratic problems that we need to solve in the maximization step of
the EM algorithm. In our experiments, we do not notice a drop in performance by
using this simplification, but we are able to increase processing speed in one
order of magnitude.
 
In summary, starting by assigning random values to the relevant parameters
$\nu_{ij}$ and $\omega_{lij}$, our EM implementation consists of
iterating the following two steps:

\begin{itemize} 
\item Expectation: estimating responsabilities for each expert using Equation
(\ref{Eq:resp}), and then estimating the outputs of gate and experts using
Equations
(\ref{Eq:MoE-Params-1}) and (\ref{Eq:MoE-Params-2}).
\item Maximization: updating the values of parameters $\nu_{ij}$ and
$\omega_{lij}$ in Equations (\ref{eq:15}) and (\ref{eq:16}) by solving
$K*(T+1)+K*(Q+1)$ quadratic problems according to the approximation described
above in Step-1 and Step-2. 
\end{itemize}

\section{Expert Selection} \label{Sec:ExpertSelection}

The MoE o RMoE assumes that all the gate functions affects to every data. But for example in object detection, we can assume that there are some group of objects i.e. group of vehicles, animals, kitchen stuff, where each group is assigned to a gate function. We think that considering all groups of objects can confuse the classifiers. Therefore we propose to select a subset of gates function according to each data. We denominate this idea as a ``expert selection''. 

Recalling that the likelihood in regular mixture of experts is:

\begin{eqnarray}\label{Eq:ExpLogLES}
	\textsl{L} & = & \prod^{N}_{n=1}{\prod^{K}_{i=1}{ p(y_n|x_n,m_i)p(m_i|x_n)}}
\end{eqnarray}

Now, in order to select a gate, we change the multinomial logit representation of the gate function (Equation \ref{Eq:MoE-Params-1}) in this way:

\begin{eqnarray} \label{Eq:CMoE-Params-1}
p(m_i|x_n) & = &  \frac{ exp{\mu_{in}(\nu^{T}_{i}x)}}{\sum^{K}_{j=1}  exp{\mu_{jn}(\nu^{T}_{j}x)}} 
\end{eqnarray}

\noindent where all the components of Equation \ref{Eq:MoE-Params-1} remain the same, except $\mu$. The variable $\mu_{in} \in \left\{0,1\right\}^K$ is the vector of model parameters of the ´´expert selector´´. It depends on data $x_n$ and expert $i$, where $i \in \{1, \dots, K\}$ for the set of expert gates. When $\mu_{in}=1/0$, it indicates that the gate $i$ is relevant/irrelevant for data $n$. In the case of $\mu_{in}=0$, the value is constant and we can say that the data $n$ is ignorant about expert $i$ and assign a constant value. In this way, it is done the expert selection.

In order to use EM algorithm, we show the expected log-likelihood by considering the \textit{responsabilities}, i.e. the posteriori probability of experts and the respective regularization terms with the addition of the term corresponding to the expert selector:

\begin{eqnarray}\label{Eq:ExpLogLES}
	\left\langle \textsl{L}_c \right\rangle & = & \sum^{N}_{n=1}{
\sum^{K}_{i=1}{ R_{in}\: \left[log \; p(y_n|x_n,m_i)\: + log \; p(m_i|x_n)\right]}}  \nonumber \\ 
&& - \lambda_\nu \sum^{K}_{i=1}{ \sum^{D}_{j=1}{ \left| \nu_{ij}
\right| }} - \lambda_\omega \sum^{Q}_{l=1}{ \sum^{K}_{i=1}{ \sum^{D}_{j=1}{
\left| \omega_{lij} \right| }}} - P(\mu) 
\end{eqnarray}

%

The penalization $P$ depends on the regularization norm, mainly 0-norm or 1-norm. Now, we define the posteriori probability of the gates $m_i$ as:

\begin{eqnarray} \label{Eq:respselexp}
R_{in} & = & \frac{p(y_n|x_n,m_i)p(m_i|x_n)}{\sum^{K}_{j=1} p(y_n |x_n,m_j)\:p(m_j | x_n)}  
\end{eqnarray}

%


Next, we repeat the strategy of Lee et al. by first optimizing the unregularized expected log-likelihood and then, adding the restriction. In order to facilitate the calculations, we define some auxiliary variables. As the derivative is linear in the sum, we calculate the contribution of a single data and call it as $E'$: 

\begin{eqnarray}\label{Eq:E_ind}
E'&=&-log\sum^{K}_{k=1}{ p(y_n|x_n,m_k)\: p(m_k|x_n)} 
\end{eqnarray}

We solve this process using an EM algorithm, where in the E-step, we calculate the responsabilities in this case by using the equation \ref{Eq:respselexp}. In the M-step, we assume the responsabilities as known and we find the optimal parameters $\nu$, $\omega$ and $\mu$.

Since the use of the responsability values, the term $p(y_n|x_n,m_k)$ can be evaluated separatevely and then the parameter $\omega$ can be optimized using the equation used in RMoE. In the case of $p(m_k|x_n)$, by fixing the parameter $\mu$, we can optimize the parameter $\nu$. 

We use some notations in order to facilitate the calculus: the term $p(y_n|x_n,m_k)\: $ as $g_k^n$, $p(m_k|x_n)$ as $h_{kn}$ and $exp(\mu_{in}\nu_{i}x_n)$ as $z_{i}$, we derive the equation respect to $\nu_{in}$ for having: 


\begin{eqnarray}\label{Eq:Total}
\pd{E'}{\nu_{i}} &=& {\pd{E'}{z_{i}}} {\pd{z_{i}}{\nu_{i}}} \nonumber \\
\pd{E'}{\nu_{i}} &=& {\left[ \sum^{K}_{k=1} {{\pd{E'}{h_{k}}}{\pd{h_{k}}{z_{i}}}} \right]} {\pd{z_{i}}{\nu_{i}}} 
\end{eqnarray}

Now we have three terms and we evaluate the derivative over each one : 

\begin{eqnarray}\label{Eq:Comp1}
\pd{E'}{h_{k}}&=&\pd{-log \sum^{K}_{j=1}{g_j h_j}}{h_k} \nonumber  \\  
\pd{E'}{h_{k}}&=&\frac{-g_k}{\sum^{K}_{j=1}{g_j h_j}} \nonumber  \\  
\pd{E'}{h_{k}}&=&-\frac{R_{kn}}{h_k} 
\end{eqnarray}

\begin{eqnarray}\label{Eq:Comp2}
\pd{h_{k}}{z_{i}}&=&\pd{\frac{exp(h_{k})}{\sum^{K}_{j=1}{exp(h_{j})}}}{z_{i}} \nonumber  \\  
\pd{h_{k}}{z_{i}}&=&\delta_{ki}h_{i}-h_{i}h_{k} 
\end{eqnarray}

\begin{eqnarray}\label{Eq:Comp3}
\pd{z_{li}}{\nu_{i}}&=&\pd{\mu_{i}\nu_{i}x}{\nu_{i}} \nonumber  \\  
\pd{z_{li}}{\nu_{i}}&=&\mu_{i}x \nonumber
\end{eqnarray}

We integrate these elements for obtain:

\begin{eqnarray}\label{Eq:newnu}
\pd{E'}{\nu_{i}} &=& \left( \sum^{K}_{k=1} {\frac{R_{kn}}{h_k} (\delta_{ki}h_{i}-h_{i}h_{k})}\right)\mu_{i}x\nonumber \\
\pd{E'}{\nu_{i}} &=& \left( R_{in} - h_{i} \right)\mu_{i}x 
\end{eqnarray}

By considering all the data, the regularization term and applying the trick of Bishop by taking the logarithms of the outputs and equaling to zero, we have:

\begin{eqnarray}
    \min_{\nu_{i}} & \sum^{N}_{n=1}{\left( (log(R_{in}) - \nu^{T}_{i} \mu_{in} x_n  \right)^2 } \nonumber\\
\mbox{subject: to}    & ||\nu_{i}||_1 \leq \lambda_\nu
     \label{eq:newnu}
\end{eqnarray}

In this case it is a modified version of equation \ref{eq:16} and we can apply a QP package to solve it. Finally, we fix the parameters $\nu$ and $\omega$ for optimizing the parameter $\mu$. The regularization over the parameter of expert selector has originally norm $0$; on the other hand, it can be relaxed bu considering norm $1$. We state both approaches:
\newline

\textbf{A. Optimization of $\mu$ considering norm $0$}

As the parameter $\mu$ depends on data $x_n$, we need to solve the optimization problem: 

\begin{eqnarray}
    \min_{\mu_{in}}& {-log\sum^{K}_{k=1}{ p(y_n|x_n,m_k)\: p(m_k|x_n)} } \nonumber \\
\mbox{subject: to}    & :||\mu_{in}||_0 \leq \lambda_\mu 
\label{eq:reg0}
\end{eqnarray}

The minimization of equation \ref{eq:reg0} requires an exploration of $C^{K}_{\lambda_\mu}$ combinations, however, by assuming a low number of gates $K<50$ and a lower number of active experts $\lambda_\mu<10$, this numerical optimization is feasible in practice. 
\newline

\textbf{B. Optimization of $\mu$ considering norm $1$}

A more applicable approach is relaxing the constraint of $0$-norm by replacing by the use of a $1$-norm, also known as LASSO regularization. Given that $\mu$ is in the same component of $\nu$, its solution has many equal steps. In particular, we find almost the same equations. Using the same notations of Equation \ref{Eq:Total}, we have for the individual log-likelihood:

\begin{eqnarray}\label{Eq:Totalmu}
\pd{E'}{\mu_{in}} &=& {\pd{E'}{z_{i}}} {\pd{z_{i}}{\mu_{in}}} \nonumber \\
\pd{E'}{\mu_{in}} &=& {\left[ \sum^{K}_{k=1} {{\pd{E'}{h_{k}}}{\pd{h_{k}}{z_{i}}}} \right]} {\pd{z_{i}}{\mu_{in}}} 
\end{eqnarray}

We get the same Equations \ref{Eq:Comp1} and \ref{Eq:Comp2}. In the case of the last component we have:

\begin{eqnarray}\label{Eq:Comp3mu}
\pd{z_{li}}{\mu_{in}}&=&\pd{\mu_{in}\nu_{i}x}{\mu_{in}} \nonumber  \\  
\pd{z_{li}}{\mu_{in}}&=&\nu_{i}x 
\end{eqnarray}

We ensemble all components equations to have:

\begin{eqnarray}\label{Eq:004}
\pd{E'}{\mu_{in}} &=& \left( \sum^{K}_{k=1} {\frac{R_{kn}}{h_k} (\delta_{ki}h_{i}-h_{i}h_{k})}\right)\nu_{i}x\nonumber \\
\pd{E'}{\mu_{in}} &=& \left( R_{in} - h_{i} \right) \nu_{i}x \nonumber 
\end{eqnarray}

In order to find the optimum parameter $\mu_{in}$, we fix $n$ and consider from $i=1$ to $K$. By equaling each equation to zero, we have: 

\begin{eqnarray}\label{Eq:mu}
\left( R_{in} - h_{i} \right) \nu_{i}x &= &0
\end{eqnarray}

Next, we approximate the previous equation using the logarithms over the outputs (Bishop):

\begin{eqnarray}\label{Eq:mulog}
\left( log(R_{in}) - \mu_{i}\nu_{i}x \right) \nu_{i}x & = &0 
\end{eqnarray}

%

Now, we fix $n$ in order to find jointly the parameters of $\mu$ for a fixed data $n$. Therefore when we add the $K$ equations, we have an equation system:

\begin{eqnarray}\label{Eq:newnu}
\left( \sum^{K}_{i=1} {\left( log(R_{in}) - \mu_{in}\nu_{i}x_n \right) \nu_{i}x_n}\right) &=&0 \nonumber \\
\end{eqnarray}

This equation can be represented as a minimization problem considering the sum of squares residuals between $log(R_{in})$ and $\mu_{in}\nu_{i}x_n$; where we add restriction of norm 1 over $\mu_{*n}$ that represents all selected experts for data $n$. In this case, we have:

\begin{eqnarray}
\min_{\mu}& {\left\| log(R_{n}) - \mu_{*n} \nu x_n \right\|_2^2} \nonumber \\
\mbox{subject: to}    & ||\mu_{*n}||_1 \leq \lambda_\mu 
\label{eq:003}
\end{eqnarray}

This equation can be solved with a quadratic program optimization package where the variable is $\mu_{*n}$. In the training phase, we require to solve this optimization $N$ times. And in the test phase, it is necessary to solve this optimization problem for each test data. 

By using norm 0 or 1, we can find the parameters of the expert selector. All the process is summarized as an EM algorithm where in the M-step, first, we freeze $\nu$ and $\omega$ and find $\mu$; then we freeze $\mu$ and iterate for finding the local optimum $\nu$ and $\omega$; then in the E-step, we find the responsabilities $R_{in}$ using the new parameters $\nu$, $\omega$ and $\mu$. In the beginning, we initialize all parameters randomly. In the following section, we will detail the results of our experiments.

\end{document}